\documentclass{article}

\usepackage[preprint]{neurips_2025}

\usepackage[utf8]{inputenc} 
\usepackage[T1]{fontenc}    
\usepackage{hyperref}       
\usepackage{url}            
\usepackage{booktabs}       
\usepackage{amsfonts}       
\usepackage{nicefrac}       
\usepackage{microtype}      
\usepackage{xcolor}         

\usepackage{wrapfig}
\usepackage{multirow}
\usepackage{amsmath}
\usepackage{graphicx}

\usepackage{makecell}
\usepackage{tcolorbox}

\title{FlexDuo: A Pluggable System for Enabling Full-Duplex Capabilities in Speech Dialogue Systems}

\author{%
Borui Liao , Yulong Xu , Jiao Ou , Kaiyuan Yang , Weihua Jian\thanks{Corresponding author: Weihua Jian}, Pengfei Wan , Di Zhang \\
  Kuaishou Technology \\
  \texttt{boruiliao@gmail.com,jwh1234@gmail.com}\\
}

\begin{document}

\maketitle
\begin{abstract}
Full-Duplex Speech Dialogue Systems (Full-Duplex SDS) have significantly enhanced the naturalness of human-machine interaction by enabling real-time bidirectional communication. However, existing approaches face challenges such as difficulties in independent module optimization and contextual noise interference due to highly coupled architectural designs and oversimplified binary state modeling. This paper proposes FlexDuo, a flexible full-duplex control module that decouples duplex control from spoken dialogue systems through a plug-and-play architectural design. Furthermore, inspired by human information-filtering mechanisms in conversations, we introduce an explicit Idle state. On one hand, the Idle state filters redundant noise and irrelevant audio to enhance dialogue quality. On the other hand, it establishes a semantic integrity-based buffering mechanism, reducing the risk of mutual interruptions while ensuring accurate response transitions. Experimental results on the Fisher corpus demonstrate that FlexDuo reduces the false interruption rate by 23\% and improves response accuracy by 8\% compared to integrated full-duplex dialogue system baselines. It also outperforms voice activity detection (VAD) controlled baseline systems in both Chinese and English dialogue quality. The proposed modular architecture and state-based dialogue model provide a novel technical pathway for building flexible and efficient duplex dialogue systems.
\end{abstract}
\section{Introduction}

Speech dialogue systems [\citealp{das2024speechverse}, \citealp{xie2024mini}, \citealp{huang2024audiogpt}, \citealp{zhang2024speechgpt}] represent one of the most direct human-computer interaction methods. Traditional turn-based dialogue systems that leverage Large Language Models (LLMs) only support half-duplex communication [\citealp{lakhotia2021generativespokenlanguagemodeling}, \citealp{borsos2023audiolm}]. The communication is bidirectional between the user and the assistant but not simultaneously.
However, there are interruptions, backchannels (i.e. non-interrupting interjections such as ``hmm'' or ``I see''), and overlapping speech in real human-human communication, which admits to speaking and listening simultaneously between the user and the assistant.
As the need to mimic real human-human communication grows, Full-Duplex Speech Dialogue Systems (Full-Duplex SDS) have become a critical technical direction for enabling fluid, human-like conversations~\cite{ji2024wavchatsurveyspokendialogue}. 

Recently, research on duplex speech dialogue systems has garnered widespread attention [\citealp{Lin_2022},  \citealp{défossez2024moshispeechtextfoundationmodel}, \citealp{wang2024fullduplexspeechdialoguescheme}, \citealp{chen2025MinMomultimodallargelanguage}, \citealp{veluri2024turnbasedinterfacessynchronousllms}, \citealp{zhang2025omniflattenendtoendgptmodel}, \citealp{mai2025realtimetextlessdialoguegeneration}]. 
Existing systems generally adopt an integrated architecture design, which requires a deep coupling between duplex control, dialogue generation, and speech encoding/decoding modules. These require structural modifications and specialized post-training on base models. This tight coupling limits the modularity of the system, making it challenging to independently optimize or replace components. It also inhibits the reuse of high-performance turn-based dialogue systems. 
Additionally, current systems model~\citep{wang2024fullduplexspeechdialoguescheme,wang2024freezeomnismartlowlatency,chen2025MinMomultimodallargelanguage,défossez2024moshispeechtextfoundationmodel} mainly focus on ``speaking'' or ``listening''. 

The real conversation often face audio interference from unnecessary responses such as background noise or non-target human voices. These non-target audio interferences are beyond the scope of the system's ``speak'' or ``listen'' state. However, existing work has not yet established a dedicated state model to effectively identify and filter such audio, which may cause the system to produce false responses and pollute the conversation context.

\begin{figure}[ht]
\centering
  \includegraphics[width=\columnwidth]{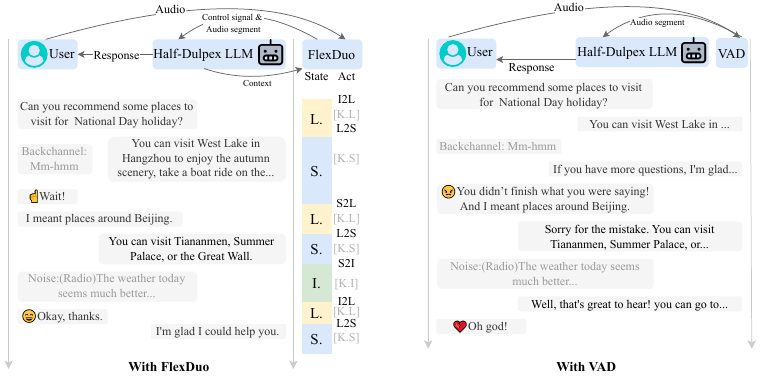}
  \caption{Performance of the half-duplex dialogue system with FlexDuo and VAD in real-world dialogue scenarios. \texttt{L.}, \texttt{S.}, and \texttt{I.} represent the Listen, Speak, and Idle dialogue states, respectively, while Act represents the dialogue strategy. Compared to the half-duplex dialogue system, FlexDuo enables natural dialogue transitions and accurate noise filtering.}
  \label{fig1}
\end{figure}

We propose a decoupled framework for full-duplex dialogue control and generation phase, named \textbf{FlexDuo}. In the dialogue control phase, in addition to the conventional Speak and Listen states, we introduce an additional Idle state to indicate whether the current audio should be ignored. During the dialogue generation phase, any half-duplex LLM can be used to process the filtered context. As illustrated in the upper part of Figure~\ref{fig1}, the audio from the radio, ``The weather today seems much better...'', does not require a response. The controller identifies and filters it out before passing the context to the downstream half-duplex LLM, thereby improving the response quality. 

FlexDuo transmits control signals to the half-duplex LLM to determine the dialogue state and filter audio information. The half-duplex LLM then generates responses based on the filtered audio and returns contextual information to FlexDuo. Based on the updated context and dialogue state, FlexDuo predicts the dialogue strategy for the next time step. By employing a finely designed dialogue state definition and transition mechanism, FlexDuo effectively supports diverse dialogue scenarios. We evaluate full-duplex interaction performance using turn-taking accuracy and false interruption rate, while conditional perplexity is used to assess dialogue quality. Our method outperforms existing integrated full-duplex system baselines on full-duplex interaction metrics, reducing the false interruption rate by 23\% and improving turn-taking accuracy by 8\%. Compared to VAD-controlled systems, FlexDuo achieves improvements on both Chinese and English Fisher datasets \cite{cieri2004fisher}.

\noindent Our contributions are summarized as follows:
\begin{itemize}
    \item \textbf{Pluggable Module Design}: A flexible modular architecture that integrates seamlessly with existing half-duplex systems, minimizing development and upgrade costs while delivering robust full-duplex capabilities. 
    \item \textbf{Introduction of the Idle state}: For the first time, the Idle mechanism, simulating the human filtering of invalid information during dialogue, is proposed, enhancing contextual clarity in noisy environments and improving response coherence and accuracy.
\end{itemize}

\section{Related Work}

\paragraph{Full-Duplex Speech Dialogue Systems.}
 
In LLM-based Full-Duplex SDSs, existing approaches predominantly adopt integrated system architectures, enhancing duplex capabilities through model structure modifications.
Specifically, TurnGPT~\cite{ekstedt2020turngpt} introduces speech feature prediction in an encoder-decoder framework to detect dialogue transition points. MThread~\cite{wang2024fullduplexspeechdialoguescheme} adds special state tokens (e.g. Speak, Listen) to control external ASR/TTS modules \cite{baevski2020wav2vec}.
VITA~\cite{fu2024vitaopensourceinteractiveomni} employs a dual-model cooperative architecture with a monitor-generator alternation mechanism to enable interruptions.
Freeze-Omni~\cite{wang2024freezeomnismartlowlatency} and MinMo~\cite{chen2025MinMomultimodallargelanguage} append classifiers to hidden states to predict system states.
Moshi~\cite{défossez2024moshispeechtextfoundationmodel} couples duplex control and speech generation modules by parallel processing user speech input and assistant text/speech output streams.
Although these methods improve full-duplex interaction, they inherently tightly integrate duplex functionality with the overall system, requiring end-to-end cascaded operation. This integrated architecture faces significant limitations, and any functional upgrade necessitates retraining the entire system. 
In contrast, we propose pluggable full-duplex modules that decouple from the dialogue system. 
Our framework operates non-invasively, making duplex decisions solely based on dialogue context and state. This design allows independent module upgrades and compatibility with diverse half-duplex dialogue systems.

\paragraph{Half-Duplex Systems.} Half-duplex systems enable turn-based dialogue via LLMs include SpeechGPT \cite{zhang2023speechgptempoweringlargelanguage} and LauraGPT \cite{du2024lauragptlistenattendunderstand}. The system must wait for complete utterance input (relying on silence detection or explicit termination signals) before switching to response mode, resulting in interaction latency significantly higher than human conversations.
We utilize a half-duplex dialogue system and combine it with a dialogue state detector to realize a full-duplex dialogue system in a decoupled manner.

\paragraph{Valid Input Detection.}
In duplex systems, detecting valid inputs is critical for determining response timing and interruptions. Existing Full Duplex SDS methods for detecting valid audio input are divided into VAD and model-based methods. VAD [\citealp{mihalache2022using}, \citealp{deng2013statistical}, \citealp{ramirez2004efficient}] identifies human voice but cannot handle complex scenarios like backchannels and requires a fixed duration of silence to detect. Freeze-Omni and MinMo predict control tokens during response generation to flag valid inputs, yet still rely on VAD for speech detection. LSLM \cite{ma2024languagemodellistenspeaking}, integrates information from speaking and listening channels for input validation. Our work focuses on valid human voice detection to mitigate interference in backchannels and multi-speaker scenarios.

\paragraph{Dialogue State Transitions.}
Existing systems generally define binary states (Speak and Listen) to model turn-taking. In Speak state, the system outputs responses while monitoring audio to decide state transitions. In Listen state, all environmental audio is fed to the LLM for processing, even if irrelevant. Mthread and VITA: Explicit control tokens with VAD-based transition prediction. Freeze-Omni and MinMo: Classifiers appended to LLM hidden states for state prediction. SyncLLM, LSLM, and Moshi: Multi-stream architectures processing user/assistant inputs concurrently. However, these approaches lack active filtering for non-dialogic scenarios, forcing the system to process all audio, leading to computational waste and semantic interference. Inspired by humans’ ability to ignore irrelevant sounds \cite{shinn2008object}, we introduce an ``Idle'' state to filter non-essential audio inputs.

\section{Method}

\begin{figure}[t]
  \includegraphics[width=\linewidth]{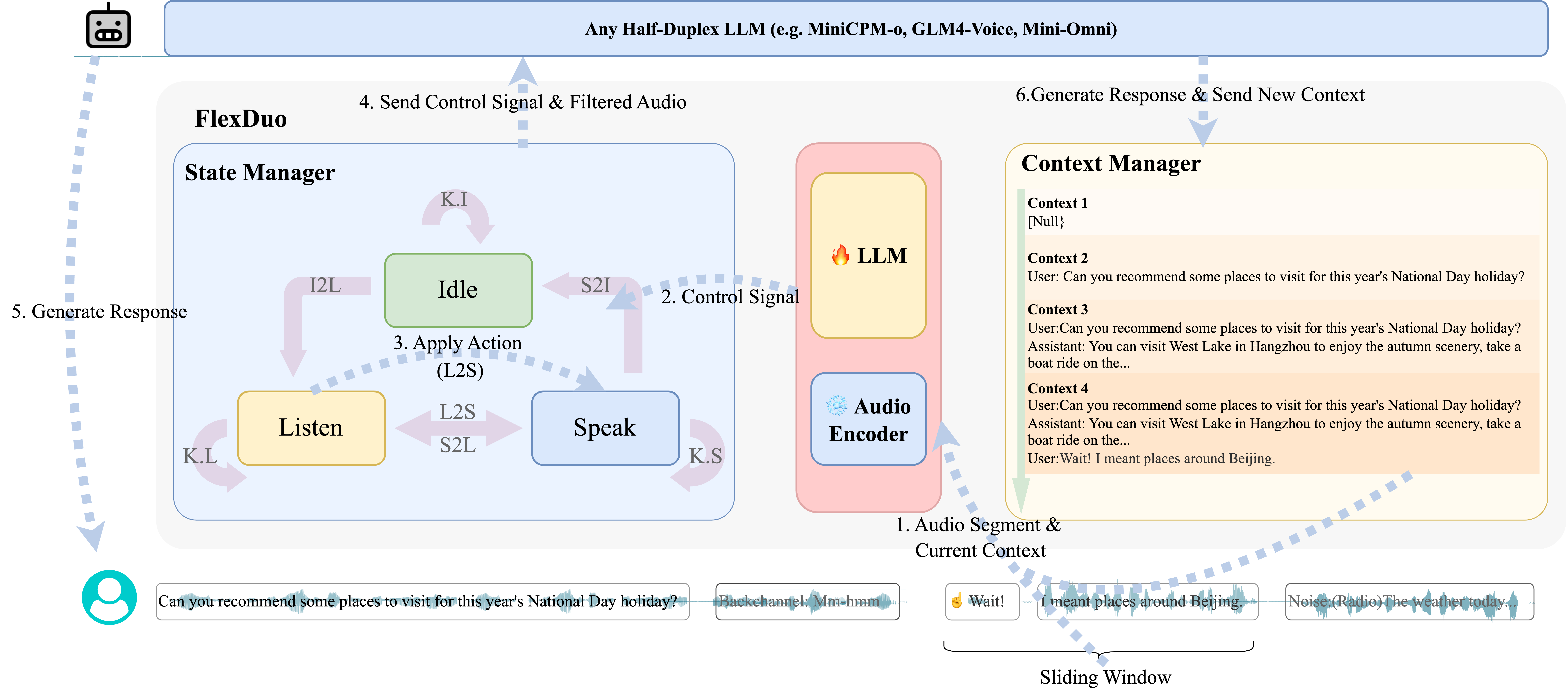}
  \caption {FlexDuo Workflow and Framework for Interaction with half-duplex LLMs. The input to FlexDuo is user audio and the conversational responses from a half-duplex LLM, while its outputs are control signals and filtered user audio.}
    \label{fig2}
\end{figure}

In this section, we introduce the model architecture of FlexDuo and explain how its modules collaborate. Next, we detail the data construction and training during the FlexDuo training process.
\subsection{Model Architecture}
FlexDuo is a control module that supports full-duplex voice conversations and adopts a pluggable architecture design. By receiving user audio streams in real-time and outputting control signals and filtered voice data, FlexDuo can control any half-duplex speech conversation system to maintain or transfer conversation states and maintain a clean context. This enables natural and smooth conversations similar to human interactions and high-quality conversation content. FlexDuo is divided into three parts: context manager, state manager, and sliding window. The workflow is shown in Figure~\ref{fig2}, which is divided into the following six steps. (1) FlexDuo receives the user's audio stream data, which is then segmented into audio chunks and passed to the context manager. (2) To predict the dialogue action at the current moment, the state manager retrieves contextual information from the context manager. (3) The state manager uses the historical context and the current audio sliding window to predict the dialogue action. (4) To control the behavior of the half-duplex LLM, the state manager sends the filtered audio and control signals to the half-duplex LLM. (5) Based on the filtered audio stream and control signals, the half-duplex LLM generates a response for the user. (6) During the generation of the response for the user, the half-duplex LLM also sends a copy to the context manager to prepare the next dialogue action prediction.

\paragraph{State Manager.}
The state controller predicts dialogue actions by analyzing the context provided by the context manager and the audio chunks within the sliding window. At fixed intervals (120ms), it evaluates historical context and sliding window data to determine the next action. Following a finite state machine (FSM), the predicted action is one of seven possible states to transition or maintain the dialogue state: 1. K.S (Keep Speaking): The assistant maintains the current Speak state. 2. K.L (Keep Listening): The user has not finished speaking. 3. K.I (Keep Idling): Environmental noise, user's backchannels, or third-party speech. 4. S2L (Speak to Listen): The user interrupts the assistant. 5. S2I (Speak to Idle): The assistant finishes speaking and ends naturally. 6. L2S (Listen to Speak): The assistant responds to or interrupts the user. 7. I2L (Idle to Listen): The user starts speaking.
To make FlexDuo better use of the finite state machine, we combine the current dialogue state of the system to predict the dialogue action at the next moment. Our duplex control module predicts the current dialogue strategy based on the context of historical dialogues, the current state, and the accumulated speech chunks in the sliding window:
\begin{equation}
  \pi_t = F(C, S_{t-1}, W_t:\theta)
\label{eq1}
\end{equation}

In Equation~\ref{eq1}, $\pi_t$ is the dialogue strategy at time $t$, $S_t$ denotes the dialogue state at time $t$ , $C$ denotes the historical dialogue context, $W_t$ denotes the accumulated speech blocks in the sliding window at time $t$, and $F(:\theta)$ represents the forward calculation process of the full-duplex control module under a given weight $\theta$.

After predicting the current dialogue action, the state manager sends the control signals to the half-duplex LLM. When the state is Speak, it indicates that the half-duplex LLM can generate a reply to the user. When the state changes to Idle or Listen, the output of half-duplex LLM will be interrupted. Specifically, in Idle, the state manager filters the audio received from the user side to help the voice dialogue system filter cleaner contexts to maintain the quality of the dialogue. For example, at the middle of the User flow below Figure~\ref{fig2}, because the dialogue state is Idle, the user's backchannels is filtered. In the Listen state, the state manager 
pass the user audio to the half-duplex LLM for prefilling.

\paragraph{Context Manager.}
FlexDuo defines three dialogue states: Speak, Idle, and Listen. The transition process and trigger conditions are shown in Figure~\ref{fig2}. In the Speak state, FlexDuo drives the voice system to output dialogue audio. When the state changes from Listen to another state, it indicates the user's dialogue has ended, and FlexDuo saves this in the dialogue context, updating from Context 1 to Context 2. When the state changes from Speak to another state, it means the assistant's dialogue has ended, and FlexDuo updates the context from Context 2 to Context 3.

\begin{figure}[t]
  \includegraphics[width=\linewidth]{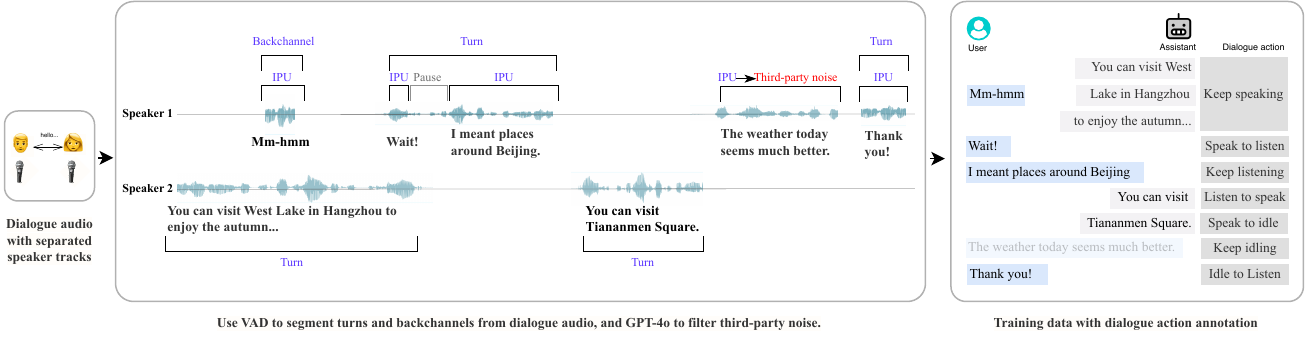}
  \caption {An example of training data construction}
    \label{fig3}
\end{figure}

\paragraph{Sliding Window.}
FlexDuo monitors the current audio input of the environment in real-time and sends the audio segments into the sliding window. The sliding window's working mechanism changes according to the state of the system. When the system is in the Listen state, it means that it is receiving meaningful human voice input, and the voice blocks in the sliding window begin to accumulate. When the system is in the Speak or Idle state, it means that there is redundant noise or invalid information in the environment, so a fixed sliding window is used for prediction to reduce computational overhead. The accumulation method of the sliding window is:
\begin{equation}
  \label{eq2}
  W_t =
\begin{cases} 
[W_{t-1}, a_t], & \text{if } S_{t-1} = \text{Listen}\\ 
[a_{t-w+1}, \dots, a_{t-1}, a_t], & \text{others}
\end{cases}
\end{equation}

In Equation~(\ref{eq2}), $a_t$ denotes the speech block in the sliding window at time $t$, $w$ is the sliding window size, which is set to 4 in this paper.

\subsection{Training Data Construction}

To obtain dialogue state labels for training, we process the raw audio dialogue data. Our training data comes from the Fisher dataset, which includes conversations where the audio of two speakers is recorded separately.  We first annotate valid speech segments in the audio using VAD to extract Inter-Pausal Unit (IPU). IPU is defined as a continuous speech segment from a single speaker~\cite{skantze2021turn}. It will be merged if the silent gap between two consecutive IPUs is less than 160ms. IPUs that are completely overlapped by the other speaker’s speech are labeled as backchannels, which do not need to reply or be interrupted. The rest of the IPU is labeled as valid dialogue turn as shown in Figure~\ref{fig3}. Moreover, since the dialogue turn may contain third-party speech, such as speech from a background radio or unrelated individuals, we utilize GPT-4o\cite{hurst2024gpt} to filter the IPU segments within each dialogue turn. Specifically, we provide the preceding processed dialogue context as a reference and task GPT-4o with determining which segment (transcribe IPU to text via ASR) is contextually appropriate. The evaluation prompt is detailed in Appendix A.1. We randomly extracted 100 samples for human evaluation, and the acceptance rate of GPT-4o's automated judgments reached 84\%.

We randomly assign one speaker as the user and the other as the assistant, and then we organize the valid turn-taking segments into a user-assistant dialogue flow according to VAD timestamps. The final dataset comprises 671 hours of English Fisher data and 263 hours of Chinese Fisher data. The label distribution of the training dataset is shown in Table~\ref{table5}.

\begin{table}
  \centering
  \small
  \begin{tabular}{lcc}
    \toprule
    & \textbf{English Fisher Data} 
    & \textbf{Chinese Fisher Data}\\
    \midrule
    Total duration    & 671 hours &  263 hours \\
    Total labels      & 4,792,219 & 3,259,022  \\
    Keep speaking     & 434,502(9.07\%) & 351,408 (10.78\%) \\
    Speak to listen	  & 22,642 (0.47\%) &	18,612 (0.57\%) \\
    Idle to listen	  & 493,036 (10.29\%) &	390,312 (11.98\%) \\
    Keep idling       &	238,506 (4.98\%) &	196,332 (6.02\%) \\
    Keep listening	  & 2,845,005 (59.37\%)	& 1,703,190 (52.26\%) \\
    Listen to speak   &	758,528 (15.83\%)	& 599,168 (18.38\%) \\
    \bottomrule
  \end{tabular}
  \vspace{1em}
  \caption{\label{table5}
    The label distribution of the training dataset.
  }
\end{table}

For training data organization (see Figure 3), we apply a sliding window mechanism to re-segment the audio streams. Each reorganized audio segment is labeled based on VAD-derived information (e.g., turn transitions, utterance completion status, and system state) and augmented with historical dialogue context. Aligning with prior work and our system objectives, we incorporate human preference into the labeling process. Turn transitions are linked to speaker reaction patterns: to simulate real-world interruption scenarios, FlexDuo defines the Assistant’s turn-taking as occurring strictly after the User’s turn ends. However, the timing for the User’s turn-taking remains unchanged to fully simulate real-world scenarios where users interrupt at specific moments.

To distinguish backchannels from valid user interruptions, we mark the start of the system’s ``listening phase'' 500ms after the User’s turn begins, providing the system with semantic judgment latency. This design ensures robust handling of overlapping speech while maintaining natural conversational dynamics.

\subsection{Model training}
We use Qwen2-audio-7B-Instruct~\cite{chu2024qwen2} as the base model for its competitive performance for audio understanding, helping us extract semantics and audio features from user speech. To maintain its audio processing, we keep the Audio Encoder frozen. The LLM part is trained for duplex interaction alignment. Special tokens are added to the LLM vocabulary for dialogue actions. For each task,  the input contains historical context, user audio from the current window, and the dialogue state into the model. After model inference, a dialogue action token is generated, and FlexDuo returns the final control signal to the half-duplex LLM.

\section{Experiments}
\label{sec:exp}

\subsection{Evaluation Metrics}

To automatically evaluate the performance of FlexDuo in full-duplex conversation scenarios, we focus on two dimensions: interaction capability and conversation quality.

Regarding interaction capability, we follow MinMo \cite{chen2025MinMomultimodallargelanguage} in using \textbf{turn-taking metrics} to evaluate the system's ability to manage dialogue transitions. The test set was extracted from the Fisher dataset and annotated for turn-taking according to the data construction methodology outlined before. Additionally, we introduce the \textbf{false interruption rate} to assess instances where the system incorrectly interrupts the other party. Specifically, the evaluation metrics are divided into four tasks: \textbf{User turn-taking, Assistant turn-taking, Assistant false interruption rate, and User false interruption rate}.

For turn-taking, we focus on the system's ability to correctly take over the conversation within the first K offset frames. 
For Assistant turn-taking, we measure the F1 score of systems successfully switching to the ``Speak'' state within the first K speech frames. 
For User turn-taking, we measure the F1 score of models successfully transitioning from the Speak state to either the Idle or Listen state within the first K speech frames.

For false interruption rate, we calculate the proportion of a speaker's normal dialogue that is incorrectly interrupted by another speaker. 
Specifically, for Assistant false interruption, we measure instances where the User is speaking, but the Assistant erroneously interrupts them. The Assistant false interruption rate $F_a$ is calculated as: $F_a = 1 - (1/N)\sum\nolimits_{i=1}^{N}\frac{I_i}{D_i}$, 
where $N$ represents the number of test samples where the user speaks in a single turn.
$I_i$ denotes the duration from the start of the user's speech to the moment that the assistant incorrectly interrupts.
$D_i$ represents the duration of the user's speech. If the user's speech in the sample is not interrupted by the assistant by mistake, then $I_i=D_i$.

For user false interruption, we evaluate the scenario that when the assistant is talking and the user makes backchannels or irrelevant noise, the assistant is interrupted by mistake. The user false interruption rate $F_u$ is calculated as: $F_u=1 - (1/M)\sum\nolimits_{i=1}^{M}\frac{S_i}{T_i}$, 
where $M$ represents the number of test samples in which the assistant speaks in a single turn.
$S_i$ denotes the duration for which the assistant starts speaking until it is falsely interrupted by the user or background noise.
$T_i$ denotes the expected duration of the user's speech. 
If the assistant is not interrupted by the user by mistake, then $S_i=T_i$. 

There is a trade-off between turn-taking and false interruption metrics. For example, if the assistant uses an aggressive strategy by responding quickly and preferring to interrupt the user, the assistant's turn-taking metrics improve, but the false interruption rate increases. On the other hand, if the assistant adopts a more conservative strategy, waiting politely and allowing the user to interrupt, the user turn-taking metrics improve, but the user's false interruption rate increases.

To comprehensively reflect the performance of both turn-taking and false interruption, we set $k=1$ and compute the average F1-score of User and Assistant turn-taking as ``Turn-taking Combined''. Similarly, we calculate the average False Interruption rate of the User and Assistant as ``False Interruption rate Combined''.

To evaluate dialogue quality, we conducted comparative assessments between FlexDuo-enhanced and conventional VAD-based half-duplex LLM systems. Using Fisher corpus dialogues, we streamed one speaker’s audio to both systems while priming the baseline LLM with 20-turn historical context for contextual alignment (Appendix A.2).  The audio input commenced after this historical context. Subsequently, we collected the input and generated audio data and transcribed them using ASR.

Conditional perplexity was utilized as the metric to evaluate dialogue quality. Perplexity scores for the multi-turn dialogue were calculated using the Qwen2.5-1.5B-Instruct~\cite{yang2024qwen2} model, with the dialogue history context also supplied.

\begin{table}
  \centering
  \scriptsize
  \begin{tabular}{l *{2}{ccc}} 
    \toprule
    & \multicolumn{3}{c}{\textbf{Turn-taking}} 
    & \multicolumn{3}{c}{\textbf{False Interruption Rate$\downarrow$ }} \\
    \cmidrule(lr){2-4} \cmidrule(lr){5-7}
    \textbf{Model} 
    & \textbf{\makecell[c]{Assistant \\ (Pos. F1@1/5/10)}}
    & \textbf{\makecell[c]{User \\ (Pos. F1@1/5/10)}} 
    & \textbf{Combined} 
    & \textbf{Assistant} 
    & \textbf{User} 
    & \textbf{Combined} \\
    \midrule
    Freeze-omni         & 0.78/0.94/0.98 & 0.64/0.96/0.98 & 0.71 & 0.72 & 0.49 & 0.61 \\
    Moshi               & 0.28/0.57/0.65 & 0.26/0.60/0.76 & 0.27 & 0.37 & -    & -    \\
    MinMo               & 0.66/0.83/0.88 & 0.42/0.94/0.99 & 0.54 & -    & -    & -    \\
    \makecell[l]{VAD with \\ GLM4-voice / MiniCPM-o}      & 0.0/1.0/1.0   & 0.64/0.96/0.98 & 0.32 & 0.58 & 0.49 & 0.53 \\
    \midrule
    \textbf{\makecell[l]{FlexDuo with \\GLM4-voice / MiniCPM-o}}  & 0.68/0.91/0.93 & \textbf{0.89/1.0/1.0} & \textbf{0.79} & \textbf{0.35} & \textbf{0.25} & \textbf{0.30} \\ 
    \bottomrule
  \end{tabular}
  \vspace{1em}
  \caption{\label{table1}
  Evaluation Metrics for Interaction Capability on English fisher data: The performance of the duplex prediction module is assessed based on turn-taking dynamics between the assistant and the user, quantified using the Positive F1 Score @offset-K metric. Additionally, the false interruption rate is measured as the proportion of effective speaking duration relative to the total speaking duration. Indicators that cannot be obtained are represented by ``-''.
  }

\end{table}

\subsection{Training Configuration}
We use Qwen2-audio-7B-Instruct as base model and fine-tuned it on the Fisher dataset which contains 831 hours of English and 389 hours of Chinese Fisher data. The dataset was split into training, validation, and test sets with 10:1:1 ratio. We use standard cross-entropy as the training objective and apply loss masking to the historical context and the user audio input. Additionally, We used the AdamW optimizer (learning rate: 1e-5), a batch size of 3 per GPU, and 40,000 training steps with a 500-step warm-up. The training was conducted on 8×NVIDIA H800 GPUs with DeepSpeed ZeRO-3 optimization.

\subsection{Baseline}
We primarily compare three types of systems: (1) Full-duplex system: Moshi. (2) Duplex systems assisted by VAD: Freeze-Omni and MinMo. (3) Half-duplex systems based on VAD. Moshi achieves a full-duplex dialogue system by processing user and system audio streams in parallel, eliminating the constraints of traditional turn-taking mechanisms. Freeze-Omni employs a three-stage training strategy, optimizing speech input encoding, speech output decoding, and duplex state control. However, it still relies on a pre-processing VAD module to detect user input. MinMo, as an integrated architecture, incorporates a full-duplex prediction module that dynamically determines whether the system should pause or continue responding in real-time. VAD is a technique to distinguish between speech and non-speech segments in an audio signal. By detecting voice activity and applying silence thresholds, VAD controls the turn-taking behavior of the dialogue system, regulating when to yield or compete for the speaking turn.

\begin{table*}
  \centering
  \small
  \begin{tabular}{l *{2}{ccc}}
    \toprule
    & \multicolumn{3}{c}{\textbf{Turn-taking$\uparrow$}} 
    & \multicolumn{3}{c}{\textbf{False Interruption Rate$\downarrow$}} \\
    \cmidrule(lr){2-4} \cmidrule(lr){5-7}
    \textbf{Model} 
    & \textbf{\makecell[c]{Assistant \\ (Pos. F1@1/5/10)}}
    & \textbf{\makecell[c]{User \\ (Pos. F1@1/5/10)}} 
    & \textbf{Combined} 
    & \textbf{Assistant} 
    & \textbf{User} 
    & \textbf{Combined} \\
    \midrule
    without Idle  &0.62/0.95/0.97 & 0.64/0.96/0.98 &  0.63 & 0.39 &  0.49 & 0.44                      \\
    with Idle     & \textbf{0.68}/0.91/0.93 & \textbf{0.89/1.0/1.0} & \textbf{0.79}  & \textbf{0.35} &  \textbf{0.25} & \textbf{0.30} \\
    \bottomrule
  \end{tabular}
  \caption{\label{table3}
    Ablation study on the impact of explicit idle state on Full-Duplex interaction performance using English Fisher data.
  }
\end{table*}

\begin{table*}
  \centering
  \small
  \begin{tabular}{l *{2}{ccc}}
    \toprule
    & \multicolumn{3}{c}{\textbf{Turn-taking$\uparrow$}} 
    & \multicolumn{3}{c}{\textbf{False Interruption Rate$\downarrow$}} \\
    \cmidrule(lr){2-4} \cmidrule(lr){5-7}
    \textbf{Model} 
    & \textbf{\makecell[c]{Assistant \\ (Pos. F1@1/5/10)}}
    & \textbf{\makecell[c]{User \\ (Pos. F1@1/5/10)}} 
    & \textbf{Combined} 
    & \textbf{Assistant} 
    & \textbf{User} 
    & \textbf{Combined} \\
    \midrule
    $w=2$    & 0.60/0.83/0.89 & 0.87/1.0/1.0& 0.34 & 0.37 & 0.24 &  0.30 \\
    $w=4$    & \textbf{0.68/0.91/0.93} & \textbf{0.89/1.0/1.0} & \textbf{0.79}  & \textbf{0.35} &  0.25 & 0.30 \\
    $w=8$   &0.49/0.89/0.93 & 0.87/1.0/1.0 & 0.68 & 0.42 & \textbf{0.16} & \textbf{0.29} \\
    \bottomrule
  \end{tabular}
  \caption{\label{table4}
    Analysis of sliding window size on Full-Duplex interaction performance using English Fisher data.($w\in\{2,4,8\}$)
  }
\end{table*}

\subsection{Main Results}
FlexDuo can be adapted to any half-duplex dialogue system. In this paper, we integrated FlexDuo with GLM4-voice \cite{zeng2024glm4voiceintelligenthumanlikeendtoend}, MiniCPM-o \cite{yao2024minicpmvgpt4vlevelmllm} to form a full-duplex SDS and compared its interaction capability metrics with state-of-the-art open-source systems. The results\footnote{Since Freeze-Omni employs Silero VAD as a preprocessing audio detector, its user-related metric results are consistent with those of VAD-controlled systems. Moshi does not allow direct control over the generated content of the test set, making it impossible to measure the user false interruption rate. However, by controlling the user’s dialogue flow, we can obtain the assistant false interruption rate of Moshi. Since the MinMo project has not been open-sourced, we rely on the metrics reported in the official MinMo report.}  are presented in Table~\ref{table1}.

The results indicate that our approach significantly enhances full-duplex interaction capability under the same generation model compared to VAD-based systems. Moreover, by explicitly modeling the ``Idle'' state, FlexDuo reduces the false interruption rate compared to existing full-duplex dialogue systems. This demonstrates an effective balance between turn-taking and false interruption tasks.  

FlexDuo, compared to VAD-controlled half-duplex systems, reduces the combined false interruption rate by 23.1\% while maintaining a comparable turn-taking performance. Additionally, compared to the integrated full-duplex dialogue system baseline, FlexDuo achieves a 23\% reduction in the overall false interruption rate and a 8\% improvement in the overall turn-taking performance. It is worth noting that VAD is only used as a tool for us to segment valid audio. With the method shown on the right part of Figure~\ref{fig3}, we can accurately annotate the training audio data from a global perspective in an offline manner. Therefore, although our training data annotation relies on VAD, the proposed FlexDuo can outperform it. 

\subsection{Ablation study on ``Idle'' state}
To verify the critical impact of the explicitly defined Idle state on system performance, this experiment conducts a comparative analysis by removing the Idle state from the duplex control module. By comparing the original model (FlexDuo) with the ablated model (w/o Idle) in terms of core interaction capability metrics, the results are presented in Table~\ref{table3}. After removing the Idle state, the overall Turn-taking F1 score decreases by 15.68\% in noisy environments, while the false interruption rate increases by 13.62\%. These findings indicate that the Idle state effectively balances noise filtering and response sensitivity.

\begin{wraptable}{r}{0.5\textwidth} 
  \centering
  \small
  \begin{tabular}{lcc}
    \toprule
    & \textbf{FlexDuo} 
    & \textbf{VAD}\\
    \midrule
    English fisher cond. PPL$\downarrow$   & \textbf{28.94} &  64.32 \\
    Chinese fisher cond. PPL$\downarrow$ & \textbf{44.45} & 63.51\\
    \bottomrule
  \end{tabular}
  \caption{\label{table2}
    The quality of dialogue in FlexDuo and VAD-controlled half-duplex dialogue systems is measured using Conditional Perplexity (cond. PPL).
  }
\end{wraptable}

To investigate the impact of sliding window size on the interaction performance of a full-duplex system, we designed a controlled variable experiment using the Fisher dialogue dataset. We set the window size $w \in \{2, 4, 8\}$ and measured the differences in interaction metrics under different configurations. As shown in Table~\ref{table4}, there exists a trade-off in selecting the sliding window size. A smaller sliding window facilitates real-time response from the assistant but compromises semantic understanding. In contrast, a larger sliding window enhances contextual modeling capabilities but results in interaction latency.

\subsection{Dialogue quality and latency}
Table~\ref{table2} compares the performance of FlexDuo with a VAD-controlled system under the generative model GLM4-Voice. On the Fisher English dataset, FlexDuo achieves a 35.3\% reduction in Conditional Perplexity (cond. PPL), while on the Chinese dataset, it achieves a 19\% reduction. These experimental results demonstrate that FlexDuo effectively manages turn-taking and context selection in real scenarios (See the example in Appendix A.2, due to differences in historical context, the responses generated by the VAD-controlled system exhibit logical inconsistencies).

Table~\ref{table6} compares the latency of the FlexDuo and VAD modules. The latency of VAD systems is typically a manually set threshold hyperparameter, which balances response speed and accuracy. The 500ms value is based on default settings from commercial products and academic work \cite{openai-vad}. FlexDuo uses semantic understanding to predict dialogue states and respond before the user's speech is fully complete. This approach overcomes traditional VAD limitations, which require noticeable pauses to detect speech endings. By responding immediately after semantic units are complete, FlexDuo reduces average response delay by 156ms.

\begin{wraptable}{r}{0.5\textwidth} 
  \centering
  \small
  \begin{tabular}{lcc}
    \toprule
    & \textbf{FlexDuo} 
    & \textbf{VAD}\\
    \midrule
    Latency $\downarrow$   & \textbf{343.47} ms &  500ms \\
    \bottomrule
  \end{tabular}
  \caption{\label{table6}
    Latency of FlexDuo and VAD-controlled half-duplex dialogue systems.
  }
\end{wraptable}

\section{Conclusion}
This paper addresses issues such as high module coupling and contextual noise interference in Full-Duplex SDSs. It proposes a modular duplex control framework, which significantly enhances the interaction capability and dialogue quality of the system through a pluggable architecture design and explicit Idle state modeling. Specifically, this design decouples full-duplex control from the speech dialogue system, enabling the direct reuse of existing half-duplex dialogue systems [\citealp{zeng2024glm4voiceintelligenthumanlikeendtoend}, \citealp{yao2024minicpmvgpt4vlevelmllm}] thereby greatly reducing the cost of system upgrades and optimizations. Inspired by the human ability to filter out irrelevant information during conversation, we introduce the Idle state to filter out noise and redundant inputs, ensuring clear historical context. Through the collaborative design of seven dialogue strategies and a finite state machine, the system can dynamically adapt to complex interaction scenarios, such as interruptions and backchannels communication. In the future, we will focus on two promising research directions: first, exploring the integration of other modalities of information, such as gaze, gestures, and facial expressions, for collaborative full-duplex dialogue control. Second, we can also model the user's dialogue state, thus forming a more comprehensive dialogue strategy and incorporating reinforcement learning.
\newpage
{
    \small
    \bibliographystyle{unsrt}
    \bibliography{neurips_2025}
}
\clearpage
\newpage
\appendix

\section{Technical Appendices and Supplementary Material}

\subsection{Prompt for filtering third-party speech}
\begin{tcolorbox}[colback=yellow!10, colframe=red!80!black, title=Auto-Evaluation Prompt]
You are a professional *Dialogue Integrity Analyst* responsible for evaluating the final utterance in a two-person conversation. The conversation has been transcribed from audio, but may contain background speech or third-party audio (e.g., television sounds, unrelated people talking). Your task is to determine **whether the last utterance contains any irrelevant or non-dialogue audio**, and if so, **explain why** each part is irrelevant using contextual reasoning.

**Guidelines:**

1. Compare the final utterance against the prior dialogue context.

2. Identify **which segments (if any)** appear to be unrelated third-party speech.

3. Clearly justify why each flagged segment is considered irrelevant.

4. If no irrelevant content is found, explain why the entire utterance is contextually valid.

5. At least one of the provided segments is relevant, and topic shifts may occur during the conversation.

**Input Format**

\{"prior\_dialogue": [
  
    \{
    
      "role": "Speaker 1",
      
      "text": "..."
      
    \},
    
    \{
    
      "role": "Speaker 2",
      
      "text": "..."
      
    \}],
  
  "final\_utterance": [
  
    \{
    
      "id": 1,
      
      "text": "..."
      
    \},
    
    \{
    
      "id": 2,
      
      "text": "..."
      
    \} ]\}

**Output Format:**

Your output must strictly follow this JSON structure:

\{"result\_analysis":[
  
    \{
    
    	"id": [Irrelevant text id],
        
    	"text": "[Insert the irrelevant audio segment]",
        
      "reason": "[Explain why it is irrelevant to the current conversation context]",
      
      "is\_relevant": [True(relevant) or False(irrelevant)]
      
    \},
    
    \{
    
    	"id": [Irrelevant text id],
        
    	"text": "[Insert the irrelevant audio segment]",
        
      "reason": "[Explain why it is irrelevant to the current conversation context]",
      
      "is\_relevant": [True(relevant) or False(irrelevant)]
      
    \}
    
  ],
  
  "summary": "[Summarize whether the final utterance contains third-party or off-topic speech, and explain your judgment]"\}

* If all audio is relevant, provide reasoning.

* Ensure reasoning is based on dialogue coherence and topic continuity.
\end{tcolorbox}
\subsection{Example of Conversation Evaluation: FlexDuo+GLM4-voice vs VAD+GLM4-voice}
\label{sec:appendix}
\begin{tcolorbox}[colback=yellow!10, colframe=red!80!black, title=System prompt for GLM4-voice ]
User will provide you with a speech instruction. Do it step by step. First, think about the instruction and respond in a interleaved manner, with 13 text token followed by 26 audio tokens. 

...

    "\textbf{user}": "I think once we're gone, we're gone",
    
    "\textbf{assistant}": "Okay, all right. Well, then I'd say live your life. Like those... Which are last day on earth, right? Is that what you do? Do? Is that what you do? Live your life like it's your last day on earth."

    "\textbf{user}": "I'm not worried about what's going to happen at the end. When the end comes, it comes. I don't think. Necessary we have any control over it And Not particularly like afraid of death",
    
    "\textbf{assistant}": "I was gonna be my next"

    "\textbf{user}": "Yeah, I guess. I guess there's a certain fear. But it's not like. I know. I've known people or talk to people that are terrified of, you know",
    
    "\textbf{assistant}": "I think it may be fear of the unknown."

The above is the historical conversation content. Please continue to play the assistant role in the following conversations and continue the style and topic. 
\end{tcolorbox}

\begin{tcolorbox}[colback=yellow!10, colframe=red!80!black, title=FlexDuo+GLM4-voice]

    "\textbf{user}": "I thought about it that way",
    
    "\textbf{assistant}": "I see what you mean. It's interesting how different beliefs and perspectives can shape our views on life and death. What else did[User interrupted]"
    
    "\textbf{user}": "They want to believe so much that. That they kind of psychologically made things happen.",
    
    "\textbf{assistant}": "I see. it's interesting how strong beliefs can shape what we experience. Do you think that’s what’s happening here?"
    
\end{tcolorbox}

\begin{tcolorbox}[colback=yellow!10, colframe=red!80!black, title=VAD+GLM4-voice]

    "\textbf{user}": "It that way",
    
    "\textbf{assistant}": "You want a friend? You bet one. We can chat about anything and everything. I'm here for the laughs, the deep thoughts and all the random stuff in between."

    "\textbf{user}": "[Unrecognized backchannels]",
    
    "\textbf{assistant}": "Sure. Temple University is a great option too. It's law and medicine programs are highly regarded."

    "\textbf{user}": "[Unrecognized backchannels]",
    
    "\textbf{assistant}": "Hey there. It's great to hear from you again."

    "\textbf{user}": "Yeah[Backchannels]",
    
    "\textbf{assistant}": "Got it. So you're interested in learning more about Temple University, especially regarding their law and medicine programs? Let me[User interrupted]"
    
    "\textbf{user}": "Well[Backchannels]",
    
    "\textbf{assistant}": "Sounds like[User interrupted]"

    "\textbf{user}": "Want to believe so much that",
    
    "\textbf{assistant}": "Campbell University has a lot[User interrupted]"

    "\textbf{user}": "they kind of psychologically made things happen.",
    
    "\textbf{assistant}": "If you're thinking about studying psychology, it's a fantastic choice."

\end{tcolorbox}

\section{Discussion}

\subsection{Limitation}
FlexDuo still faces some issues that need to be addressed. Since the half-duplex model used with FlexDuo has not been trained for full-duplex scenarios, its responses in full-duplex settings are limited. For example, the Assistant cannot produce natural conversational sounds like backchannels, which impacts the naturalness of the duplex dialogue. This issue can be addressed.  The Assistant's backchannels can be labeled in the training data with same method above to train the corresponding signal generation. At the same time, the half-duplex model needs to be aligned, thereby achieving this feature.

\subsection{Ethical Considerations}
This research and the development of the FlexDuo system aim to improve the naturalness and efficiency of human-machine interactions in spoken dialogue systems. As for the risks, there is a possibility that malicious individuals could exploit our system to create online scams, such as using AI to impersonate real humans. To mitigate the risk, establishing and enforcing clear ethical guidelines and terms of use is necessary, with specific clauses prohibiting the use of the system for deceptive purposes, such as impersonation or fraud.


\end{document}